\documentclass[journal]{IEEEtran}
\usepackage{amsmath,amssymb,booktabs,graphicx,tabularx,float,url,cite}
\usepackage{stfloats}

\begin{document}

\title{DisasterNexus: Bridging Multimodal Semantics and High-Resolution Bitemporal Representations for Unified Disaster Understanding}

\author{Zheng~Luo, Zhenfeng~Shao, Hong~Shu, Bingwei~Hui, and~Bo~Liu%
\thanks{This work was supported by the Yunnan Provincial Key Research and Development Program Project (Grant No. 202603AG380001), the Yunnan Satellite Remote Sensing Intelligent Application Key Technology and Integrated Service Platform Project for Regional Security in South and Southeast Asia (Grant No. 202403ZC380001), and the Fundamental Research Fund Program of LIESMARS (Grant No. 4201-420100071). \textit{(Corresponding author: Zhenfeng Shao.)}}%
\thanks{Zheng Luo, Zhenfeng Shao, Hong Shu, and Bo Liu are with the State Key Laboratory of Information Engineering in Surveying, Mapping and Remote Sensing, Wuhan University, Wuhan 430079, China (e-mail: shaozhenfeng@whu.edu.cn).}%
\thanks{Bingwei Hui is with the Key Laboratory of Automatic Target Recognition, School of Electronic Science, National University of Defense Technology, 410073, China.}}

\markboth{PREPRINT}%
{Luo \MakeLowercase{\textit{et al.}}: DisasterNexus: Bridging Multimodal Semantics and High-Resolution Bitemporal Representations}

\maketitle

\begin{abstract}
Unified disaster understanding requires interpreting events from pre- and post-event observations and localizing instruction-specified affected regions. While semantic interpretation relies on scene-level context, fine-grained localization depends on clear object boundaries and bitemporal spatial correspondence—a challenge further compounded by the imaging differences between optical and SAR data. To address these distinct needs, we propose DisasterNexus, a unified framework built on complementary semantic and spatial pathways. A vision--language branch jointly encodes images and instructions to support recognition, counting, relational reasoning, and text generation. In parallel, an independent high-resolution spatial branch extracts multiscale pre- and post-event features, applying sensor-conditioned adaptation and temporal fusion to preserve fine-grained spatial details. From these features, the spatial branch generates candidate masks, and a segmentation representation from the language model scores their relevance to the instruction to select the final target region. Across the nine DisasterM3 tasks, a single 4B model achieves 72.58\% macro-averaged exact-match accuracy on six short-answer tasks and 49.06\% cumulative intersection-over-union (cIoU) on referring segmentation. Under identical training data and budgets, adding the independent high-resolution spatial branch outperforms the Qwen-feature baseline by 12.70 percentage points in cIoU, while maintaining nearly identical accuracy on the six short-answer tasks. The source code is available at \url{https://github.com/MLS2021/DisasterNexus}.
\end{abstract}


\section{Introduction}

\IEEEPARstart{D}{isaster} interpretation from remote sensing must establish both what happened and where its impacts occurred. Recognizing disaster types, identifying affected facilities, and generating reports provide scene-level information for response planning, whereas delineating damaged buildings or inundated roads requires pixel-level spatial evidence. Recognizing that a road is affected, for example, does not determine the boundaries of the affected segments. DisasterM3 organizes recognition, damage quantification, relational reasoning, referring segmentation, and text generation into nine tasks \cite{wang2025disasterm3}, offering a benchmark to evaluate how a single model handles both semantic interpretation and spatial localization.

These two output types place different demands on visual representations. Semantic answering depends on contextual relationships between scene content and instructions, whereas referring segmentation must distinguish adjacent objects, preserve target boundaries, and compare pre- and post-event states. Variations in object scale, dense layouts, and damaged appearances further complicate localization in disaster imagery. Pre-event images provide structural references for object locations, and post-event images capture current conditions. With post-event SAR observations, differences in imaging mechanisms mean that appearance changes cannot be equated directly with disaster damage. BRIGHT provides a benchmark for post-event SAR building-damage assessment with pre-event optical references \cite{chen2025bright}. A unified model therefore needs contextual representations for semantic interpretation alongside fine-grained spatial representations for bitemporal comparison.

SegEarth-R2 uses hierarchical visual features and language-conditioned prediction for remote-sensing segmentation, while TerraScope explores geospatial reasoning grounded in pixel evidence \cite{xin2026segearthr2,shu2026terrascope}. Building on these studies, we investigate how to combine contextual semantics with bitemporal spatial information so that a single model can answer questions about disaster impacts and accurately delineate instruction-specified affected regions. We evaluate pixel-level localization and semantic-task performance separately to assess how well the model supports both types of output.

We propose DisasterNexus, which processes disaster imagery through complementary vision--language and high-resolution spatial paths. The vision--language path jointly encodes images and natural-language instructions to provide contextual semantics for recognition and text-based answering. In parallel, the spatial path uses a shared encoder to extract multiscale pre- and post-event features, followed by sensor-conditioned adaptation and temporal fusion for pixel prediction. The spatial path generates candidate masks from multiscale features, and the language model's segmentation representation scores their relevance to select the region corresponding to the user's request. By separating these two pathways within one instruction-based framework, DisasterNexus handles post-event optical or SAR observations across both semantic and dense-prediction tasks.

We evaluate this design across all nine DisasterM3 tasks. A single 4B model achieves 49.06\% segmentation cIoU while supporting semantic answering and long-form generation. Under identical sample sequences and task exposures, a paired full-training comparison between the Qwen-feature baseline and our independent high-resolution spatial configuration shows that the latter achieves a 12.70-percentage-point higher cIoU—driven primarily by overlap on non-empty targets—while macro-averaged accuracy across the six short-answer tasks remains nearly identical. In a separate experiment, using both pre- and post-event images in the spatial path yields higher segmentation cIoU than a post-event-only spatial configuration, with a larger gain for optical--SAR pairs. Together, these results support preserving complementary semantic and bitemporal spatial features within a unified disaster model. Our contributions are as follows:

\begin{enumerate}
\item \textbf{DisasterNexus, a unified framework for multi-task disaster understanding.} One model supports recognition, damage quantification, relational reasoning, referring segmentation, and text generation with post-event optical or SAR observations, bringing semantic answering and pixel localization into a common instruction-based interface.
\item \textbf{Complementary semantic and bitemporal spatial representations for instruction-based localization.} A vision--language path provides contextual semantics, while an independent high-resolution spatial path extracts multiscale pre- and post-event features and combines them through sensor-conditioned adaptation and temporal fusion. The spatial path generates candidate masks, and the language model's segmentation representation scores their relevance to the instruction and selects the final region.
\item \textbf{Nine-task validation of the representation design and analysis of localization bottlenecks.} Paired full-training comparisons show that the independent high-resolution spatial configuration improves cIoU by 12.70 percentage points over the Qwen-feature baseline while maintaining comparable short-answer accuracy. We further compare post-event-only and bitemporal fusion configurations and analyze candidate coverage and instruction-dependent selection. The source code is available at \url{https://github.com/MLS2021/DisasterNexus}.
\end{enumerate}

\section{Related Work}
\label{sec:related-work}

\subsection{Bitemporal and Cross-Sensor Fusion for Disaster Assessment}

Bitemporal disaster assessment combines spatial references from pre-event imagery with evidence of post-event conditions. Early change-detection architectures, including FC-Siam, use shared Siamese encoders with feature concatenation or differencing \cite{daudt2018siamese}. For building damage, DamFormer couples localization with damage estimation \cite{chen2022damformer}, while ChangeMamba models spatiotemporal dependencies and uses pre-event features for localization alongside bitemporal features for damage prediction \cite{chen2024changemamba}. Together, these designs highlight the distinct roles of locating exposed objects and interpreting their post-event states, motivating the preservation of temporal spatial features in instruction-guided prediction.

Cross-sensor observations require additional attention to feature comparability, spatial correspondence, and structural detail. BRIGHT provides pre-event optical and post-event SAR observations for all-weather building-damage assessment \cite{chen2025bright}. HeteCD addresses heterogeneous representations through feature-consistency alignment \cite{jing2025hetecd}, while M$^2$CD combines expert routing with a training-only intermediate pathway \cite{liu2025m2cd}. GLCD-DA uses image translation and global--local interactions to reduce modality differences and model changes \cite{li2025glcdda}. Other methods focus on complementary spatial cues: IM4CD examines local image-matching displacements \cite{chen2026im4cd}, M3ICNet preserves information across observations with different resolutions \cite{zhang2025m3icnet}, and SSFCNet combines spatial--frequency learning with inverse wavelet decoding for fine-grained change recovery \cite{yang2026ssfcnet}. Cross-sensor temporal comparison therefore calls for feature alignment, spatial correspondence, and sensitivity to fine-grained changes.

Language-conditioned approaches further allow users to specify the changes of interest. DynamicEarth studies open-vocabulary change detection \cite{li2025dynamicearth}, while RCD formulates text-conditioned segmentation of category-specific changes \cite{korkmaz2026rcd}, linking temporal evidence with semantic selection beyond fixed-category damage maps. For a unified disaster model, this literature underscores the importance of preserving pre-event references and post-event spatial detail within the pixel-prediction path when supporting broader semantic requests.

\subsection{Remote-Sensing Multimodal Large Language Models and Unified Disaster Understanding}

Remote-sensing multimodal large language models adapt image interpretation to a common language interface, differing in training data, observation types, and output granularity. RSGPT emphasizes high-quality remote-sensing image--text data \cite{hu2025rsgpt}, while GeoChat extends dialogue to regional grounding \cite{kuckreja2024geochat}. EarthGPT incorporates multisensor information \cite{zhang2024earthgpt}, LHRS-Bot introduces geographic information for semantic learning \cite{muhtar2024lhrsbot}, and SkyEyeGPT unifies vision--language tasks through instruction tuning \cite{zhan2025skyeyegpt}. Collectively, they establish language-based interaction and domain adaptation for diverse remote-sensing requests.

Temporal and disaster-oriented models extend the observations and tasks covered by this interface. TEOChat supports conversations about temporal Earth observations, including building change and damage assessment \cite{irvin2025teochat}; EarthDial incorporates multisensor, multitemporal, and multiresolution imagery \cite{soni2025earthdial}. DisasterM3 organizes nine tasks covering disaster recognition, damage assessment, relational reasoning, referring segmentation, and response-oriented generation \cite{wang2025disasterm3}. ChangeQuery combines pre-event optical and post-event SAR observations with structured semantic annotations for damage quantification, regional descriptions, and disaster summaries \cite{sun2026changequery}.

Scene recognition and disaster-report generation rely on contextual semantics, whereas localizing regions specified by instructions requires retaining target locations, scales, and boundaries. DisasterNexus uses complementary semantic and spatial paths to support both types of output within a single model.

\subsection{Language-Guided Dense Prediction in Multimodal Large Language Models}

Language-guided dense prediction connects textual intent with pixel-level outputs. LISA uses a segmentation token to support reasoning segmentation under implicit requests \cite{lai2024lisa}. PixelLM extends pixel reasoning to multiple targets \cite{ren2024pixellm}, while GLaMM associates generated text with segmentation masks \cite{rasheed2024glamm}. GSVA addresses multiple and absent targets \cite{xia2024gsva}, and PSALM combines multiscale visual features, mask queries, and a flexible input schema across segmentation tasks \cite{zhang2024psalm}. These studies provide established mechanisms for conditioning dense predictions on language, including semantic tokens, dedicated spatial features, and mask decoders.

Remote-sensing studies adapt these connections to scale variation, arbitrary orientations, and densely distributed objects. RRSIS establishes referring segmentation for remote sensing \cite{yuan2024rrsis}, RMSIN models cross-modal interactions across scales and orientations \cite{liu2024rmsin}, and RSRefSeg~2 separates coarse localization from refinement through foundation-model cooperation \cite{chen2026rsrefseg2}. MM-OVSeg uses optical--SAR complementarity for open-vocabulary land-cover segmentation \cite{wei2026mmovseg}. Among generative models, GeoPixel supports pixel-grounded conversations over high-resolution imagery \cite{shabbir2025geopixel}, LISAt addresses complex implicit requests \cite{quenum2025lisat}, and SegEarth-R2 broadens language-guided segmentation across hierarchical granularity, multiple targets, and reasoning requirements \cite{xin2026segearthr2}. TerraScope further investigates pixel-grounded geospatial reasoning involving multimodal and temporal observations \cite{shu2026terrascope}.

See, Say, and Segment reports that segmentation fine-tuning can weaken object-existence judgments and natural interaction, and examines the role of joint training \cite{wu2024seesaysegment}. We therefore evaluate semantic outputs and pixel-level localization separately. Building on SegEarth-R2 for segmentation, DisasterNexus generates candidate masks through the high-resolution spatial path and uses the language model's segmentation representation to score their relevance to the instruction and select the final region.

\section{Methodology}
\label{sec:methodology}

\subsection{Problem Formulation and Framework Overview}

Given remote-sensing observations $\mathcal I$, an instruction $u$, and its task type $t$, DisasterNexus returns either a semantic answer or an instruction-specified binary mask. For paired observations, $\mathcal I=\{I_{\mathrm{pre}},I_{\mathrm{post}}\}$ contains a pre-event optical image and a post-event optical or SAR image. Single-image requests are processed directly without bitemporal pairing. The post-event modality $m\in\{\mathrm{opt},\mathrm{sar}\}$ is supplied by the dataset metadata.

The model organizes visual evidence into two pathways. A vision--language path produces contextual representations for recognition and text-based answers, while an independent high-resolution spatial path preserves multiscale features for pixel prediction. Both paths process the observations associated with the instruction, using separate visual encoders tailored to their respective outputs; within the spatial path, a single shared encoder processes both time points. Across the nine tasks, DisasterNexus employs three prediction mechanisms: task-conditioned option scoring for three recognition tasks, autoregressive generation for five semantic tasks, and query-based mask prediction with instruction-conditioned selection for referring segmentation.

For segmentation, the language model supplies a representation of a special \texttt{[SEG]} token to score the relevance of spatial mask candidates to the instruction, whereas recognition and text generation use the vision--language path directly. This design bridges language understanding and pixel-level output via candidate selection while keeping task-appropriate prediction heads in one model.

\begin{figure*}[!t]
\centering
\includegraphics[width=\textwidth]{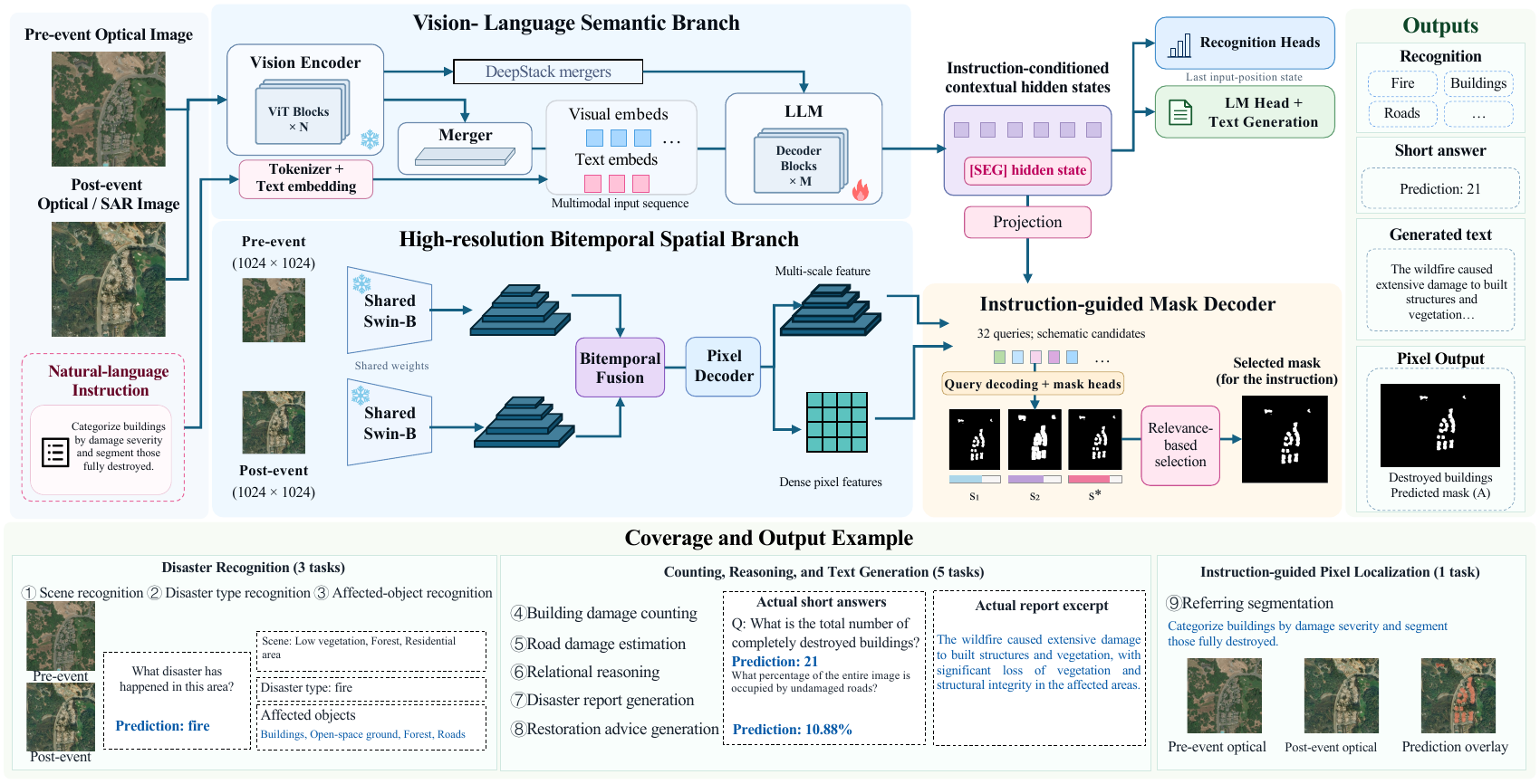}
\caption{Overview of DisasterNexus and its task coverage. The vision--language branch provides contextual representations for recognition and text generation, while the independent high-resolution bitemporal spatial branch provides dense prediction features. The \texttt{[SEG]} representation scores instruction relevance for candidate-mask selection. The lower panel illustrates recognition, counting, reasoning, text generation, and referring segmentation; the displayed inputs and DisasterNexus outputs use the same Santa Rosa wildfire case. Short answers are option-decoded and the report is excerpted verbatim. Candidate masks and relevance scores are schematic.}
\label{fig:overall-framework}
\end{figure*}

Figure~\ref{fig:overall-framework} summarizes the overall framework and the supported task outputs.

\subsection{Complementary Semantic and Spatial Representations}

\begin{figure*}[!t]
\centering
\includegraphics[width=\textwidth]{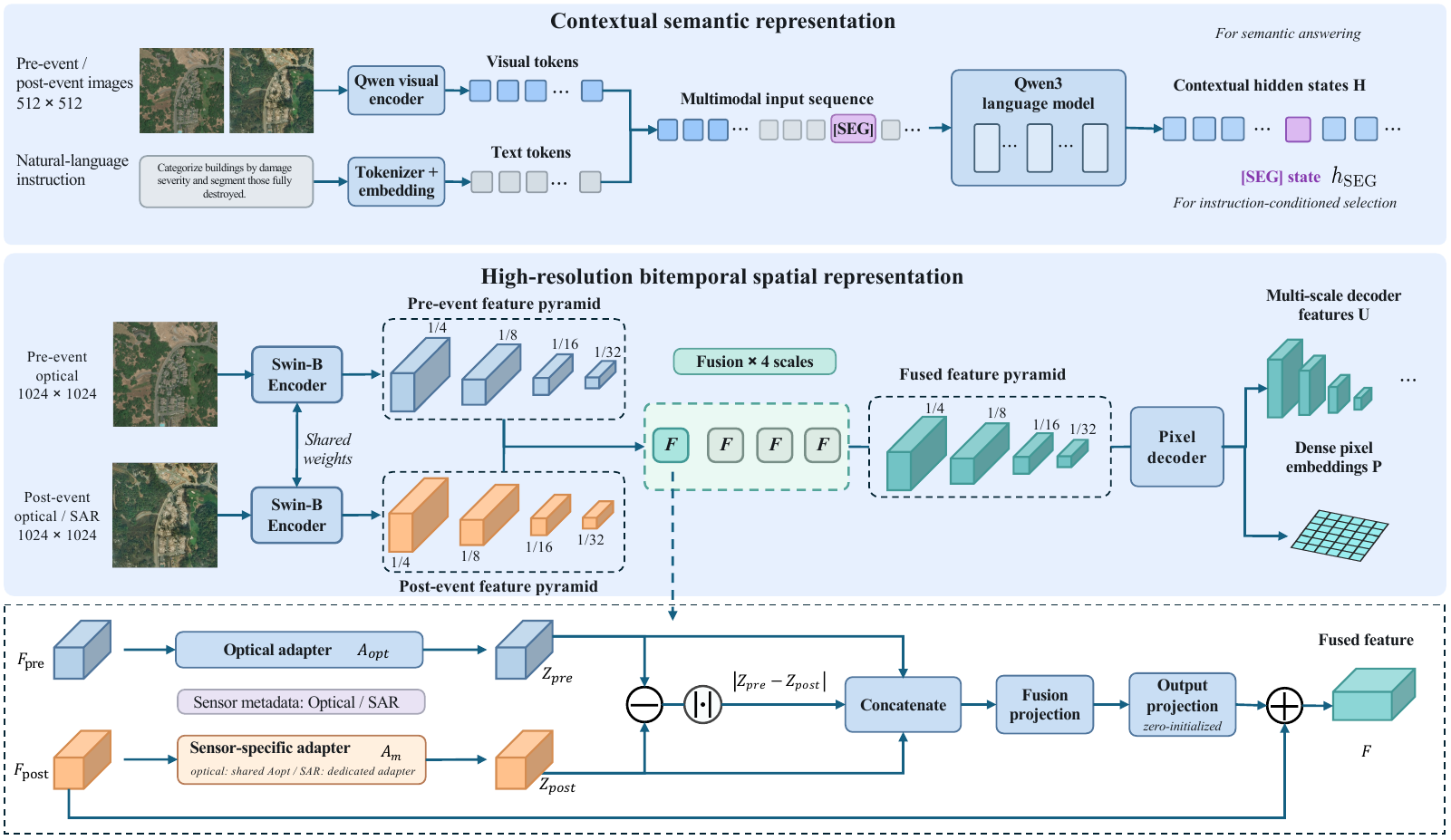}
\caption{Complementary semantic and spatial representations. Top: image and instruction tokens form contextual hidden states, including the \texttt{[SEG]} state. Middle: a shared Swin-B encoder extracts pre- and post-event feature pyramids, which are fused at four scales and passed to a pixel decoder. Bottom: sensor-specific adaptation, concatenation of adapted features and their absolute difference, and a zero-initialized output projection produce a residual update to the post-event feature.}
\label{fig:semantic-spatial}
\end{figure*}

Figure~\ref{fig:semantic-spatial} details the contextual semantic path, the high-resolution bitemporal spatial path, and sensor-conditioned fusion at one feature scale.

The semantic path uses the visual encoder $E_v$ and language model $f_\theta$ of Qwen3-VL. Images are preprocessed by $\mathcal R_v$ and combined with the tokenized instruction and any supplied answer options:
\begin{equation}
\label{eq:method-1}
\mathbf H=f_\theta\!\left(E_v(\mathcal R_v(\mathcal I)),\operatorname{Tok}(u)\right).
\end{equation}
The hidden states $\mathbf H$ provide image-conditioned language context for the task heads. For recognition, a state preceding the answer is read; for autoregressive tasks, the model predicts successive answer tokens. A segmentation request additionally uses the hidden state associated with \texttt{[SEG]}. The semantic path thus provides the contextual interpretation shared by the different output forms.

The spatial path uses a separate Swin-B encoder $E_s$ to preserve a hierarchy of spatial features:
\begin{equation}
\label{eq:method-2}
\begin{aligned}
\{\mathbf F_{\mathrm{pre}}^l\}_{l=1}^{4}&=E_s(\mathcal R_s(I_{\mathrm{pre}})),\\
\{\mathbf F_{\mathrm{post}}^l\}_{l=1}^{4}&=E_s(\mathcal R_s(I_{\mathrm{post}})).
\end{aligned}
\end{equation}
The preprocessing $\mathcal R_s$ uses a higher image resolution than $\mathcal R_v$. The four spatial levels have strides $4$, $8$, $16$, and $32$, with $128$, $256$, $512$, and $1024$ channels. Sharing $E_s$ across time points gives both observations the same feature hierarchy, independent of the Qwen visual encoder $E_v$; pre-event features provide spatial references, while post-event features reflect the observed conditions.

The adopted sensor-conditioned hierarchical fusion (SCHF) component adapts and combines this feature pyramid. At level $l$, $A_{\mathrm{opt}}^l$ is shared by optical observations, and $A_{\mathrm{sar}}^l$ has separate parameters for SAR observations. The fusion is
\begin{equation}
\label{eq:method-3}
\begin{aligned}
\mathbf Z_{\mathrm{pre}}^l&=A_{\mathrm{opt}}^l(\mathbf F_{\mathrm{pre}}^l),\quad
\mathbf Z_{\mathrm{post}}^l=A_m^l(\mathbf F_{\mathrm{post}}^l),\\
\mathbf C^l&=[\mathbf Z_{\mathrm{pre}}^l;\mathbf Z_{\mathrm{post}}^l;
|\mathbf Z_{\mathrm{post}}^l-\mathbf Z_{\mathrm{pre}}^l|],\\
\overline{\mathbf F}^{\,l}&=\mathbf F_{\mathrm{post}}^l+
P_{\mathrm{out}}^l\!\left(\Phi^l(\mathbf C^l)\right).
\end{aligned}
\end{equation}
where $[\cdot;\cdot]$ denotes channel-wise concatenation. Each adapter contains a $1\times1$ projection to 128 channels and a spatial residual operation comprising group normalization, GELU, depthwise $3\times3$ convolution, and pointwise convolution. Its channel-wise residual scale is initialized to zero. The fusion mapping $\Phi^l$ is a $1\times1$ convolution followed by group normalization and GELU; $P_{\mathrm{out}}^l$ restores the original channel width and is initialized to zero, so the initial decoder input equals the original post-event feature. The absolute difference provides a comparison cue across adapted observations, and sensor metadata select the optical or SAR adapter.

A multiscale pixel decoder transforms the fused pyramid into query-decoding features $\{\mathbf U^l\}$ and a dense pixel embedding $\mathbf P$:
\begin{equation}
\label{eq:method-4}
(\{\mathbf U^l\},\mathbf P)=\mathcal D_{\mathrm{pix}}(
\{\overline{\mathbf F}^{\,l}\}_{l=1}^{4}).
\end{equation}
The fused hierarchical features preserve spatial information at different scales and provide the input for subsequent candidate-mask prediction.

\subsection{Instruction-Conditioned Task Prediction}

Disaster type recognition (DTR), disaster scene recognition (DSR), and disaster-bearing body recognition (BBR) use task-conditioned option heads. Let $\mathbf h_u$ be the last input hidden state preceding the answer, and let $c_k$ be an option containing $|c_k|$ tokens. Its embedding and score for task $t$ are
\begin{equation}
\label{eq:method-5}
\begin{aligned}
\mathbf e_k&=\frac{1}{|c_k|}\sum_{j=1}^{|c_k|}E_{\mathrm{tok}}(c_{k,j}),\\
a_{t,k}&=\frac{\exp(\gamma_t)}{\sqrt{d_o}}\,
\phi_t(\operatorname{LN}(\mathbf h_u))^{\mathsf T}
\psi_t(\operatorname{LN}(\mathbf e_k))+b_t.
\end{aligned}
\end{equation}
$E_{\mathrm{tok}}$ is the frozen input-token embedding, $\phi_t$ and $\psi_t$ are task-specific affine projections to $d_o=256$ dimensions, and $\gamma_t$ and $b_t$ are learned scalars. The projected vectors are compared by a scaled dot product. DTR selects $\arg\max_k a_{\mathrm{DTR},k}$; DSR and BBR keep each option with $\sigma(a_{t,k})\geq0.5$, returning an empty set if none exceeds the threshold. The heads read image and instruction context before the answer tokens.

The five other semantic tasks use autoregressive prediction:
\begin{equation}
\label{eq:method-6}
p_\theta(\mathbf y\mid\mathcal I,u)=
\prod_{j=1}^{|\mathbf y|}p_\theta(y_j\mid\mathcal I,u,\mathbf y_{<j}).
\end{equation}
For bearing-body counting (BDC), disaster ratio estimation (DRE), and object relation reasoning (ORR), generated option letters are mapped to the benchmark's quantity, proportion, or relation choices. Disaster report and recovery advice generation produce free-form text. All five use the vision--language context directly.

\begin{figure*}[!t]
\centering
\includegraphics[width=\textwidth]{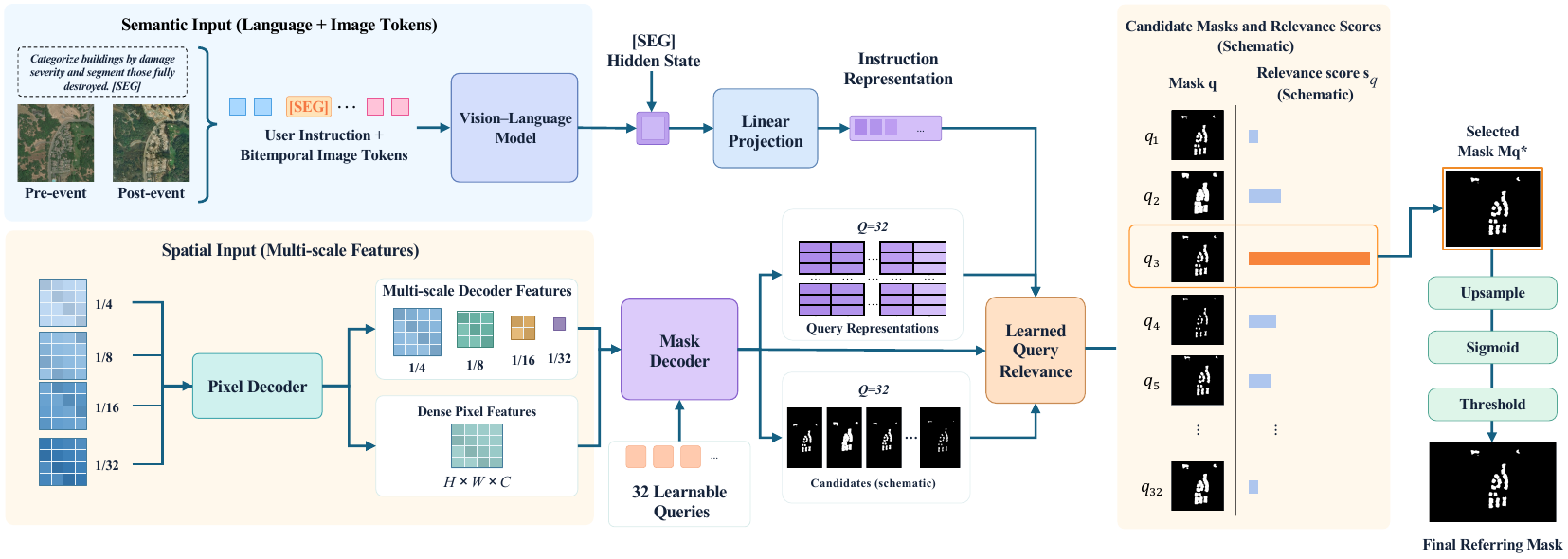}
\caption{Instruction-guided mask prediction and selection. The pixel decoder transforms the fused bitemporal spatial pyramid into multiscale decoder features and dense pixel embeddings.}
\label{fig:mask-decoder}
\end{figure*}

Figure~\ref{fig:mask-decoder} illustrates the separation between spatial candidate generation and instruction-conditioned relevance scoring. Its multiscale spatial input is the fused bitemporal pyramid produced by the preceding spatial path.

For referring segmentation, the final-layer hidden state of the \texttt{[SEG]} token is projected to a 256-dimensional instruction-conditioned representation. In parallel, a query decoder produces candidate features from the spatial pyramid and $Q=32$ learned queries $\mathcal Q$:
\begin{equation}
\label{eq:method-7}
\begin{aligned}
\mathbf z_{\mathrm{SEG}}&=\mathbf W_{\mathrm{SEG}}\mathbf h_{\mathrm{SEG}}+
\mathbf b_{\mathrm{SEG}},\\
\{\mathbf d_q\}_{q=1}^{Q}&=\mathcal D_{\mathrm{query}}(\{\mathbf U^l\},\mathbf P,\mathcal Q).
\end{aligned}
\end{equation}
$\mathbf d_q$ denotes a decoder-normalized query feature. Separate learned projections produce a mask embedding and a relevance embedding for each query:
\begin{equation}
\label{eq:method-8}
\begin{aligned}
\mathbf L_q(x)&=\phi_{\mathrm{mask}}(\mathbf d_q)^{\mathsf T}\mathbf P(x),\\
s_q&=\phi_{\mathrm{sel}}(\mathbf d_q)^{\mathsf T}\mathbf z_{\mathrm{SEG}}.
\end{aligned}
\end{equation}
$\mathbf L_q(x)$ is the mask logit at pixel $x$, and $s_q$ measures compatibility with the current instruction. The mask projection is a three-layer MLP, and the relevance projection is a two-layer MLP. Spatial features and learned queries generate the mask candidates; the segmentation-token representation scores their relevance, linking the current instruction to a candidate region.

Inference selects the highest-scoring candidate:
\begin{equation}
\label{eq:method-9}
q^*=\arg\max_q s_q,\qquad
\widehat{\mathbf M}=\mathbf 1\!\left[
\sigma(\operatorname{Up}(\mathbf L_{q^*}))\geq\tau_s\right].
\end{equation}
$\operatorname{Up}$ bilinearly resizes logits to the output grid; $\tau_s$ is fixed by the evaluation protocol. Empty-target requests follow the same prediction path. Ground-truth matching supervises candidate masks and their relevance scores during training, while inference selects a candidate using its relevance score.

\subsection{Training Objectives and Model Adaptation}

The joint-training objective combines answer-token, recognition, and mask supervision. For a mixed batch of $B$ samples, $\Omega_i$ denotes the supervised answer-token positions of sample $i$, and $N_i=|\Omega_i|$. The text loss is
\begin{equation}
\label{eq:method-10}
\mathcal L_{\mathrm{text}}=-\frac{1}{B}\sum_{i=1}^{B}\frac{1}{N_i}
\sum_{j\in\Omega_i}\log p_\theta(y_{i,j}\mid\mathcal I_i,u_i,\mathbf y_{i,<j}).
\end{equation}
Only answer positions are supervised. This term applies to all nine tasks, including the segmentation token \texttt{[SEG]}. Recognition additionally uses cross-entropy for DTR and binary cross-entropy averaged over valid options for DSR and BBR; averaging over the $n_r$ recognition samples gives $\mathcal L_{\mathrm{rec}}$.

For segmentation, $\mathbf Y_i$ is the binary union of the regions specified by the current instruction. At supervised decoder output $r$, nearest-neighbor interpolation gives $\mathbf Y_i^{(r)}$ on the mask-logit grid. With $\mathbf p_{i,q}^{(r)}=\sigma(\mathbf L_{i,q}^{(r)})$, the matched query is
\begin{equation}
\label{eq:method-11}
q_{i,\mathrm{gt}}^{(r)}=\arg\min_q\left[
\operatorname{BCE}(\mathbf p_{i,q}^{(r)},\mathbf Y_i^{(r)})+
\operatorname{Dice}(\mathbf p_{i,q}^{(r)},\mathbf Y_i^{(r)})\right].
\end{equation}
BCE is averaged over pixels. Dice loss is $1-(2\sum_jp_jY_j+\varepsilon)/(\sum_jp_j+\sum_jY_j+\varepsilon)$, with $\varepsilon=10^{-6}$. Matching uses the unweighted sum of these losses independently at each supervised output. The mask objective combines the matched mask losses with query-selection cross-entropy:
\begin{equation}
\label{eq:method-12}
\begin{aligned}
\mathcal L_{\mathrm{mask}}=\frac{1}{n_s}
\sum_{i\in\mathcal B_s}\sum_{r\in\mathcal R}\biggl[&
\lambda_b\operatorname{BCE}(\mathbf p_{i,q_{i,\mathrm{gt}}^{(r)}}^{(r)},\mathbf Y_i^{(r)})\\
&+\lambda_d\operatorname{Dice}(\mathbf p_{i,q_{i,\mathrm{gt}}^{(r)}}^{(r)},\mathbf Y_i^{(r)})\\
&+\lambda_q\operatorname{CE}(\mathbf s_i^{(r)}/T,q_{i,\mathrm{gt}}^{(r)})\biggr].
\end{aligned}
\end{equation}
Here, $\mathcal B_s$ contains $n_s$ segmentation samples, and $\mathcal R$ comprises the final output and the last three auxiliary outputs; their losses are summed. Empty-target instructions are supervised with an all-zero mask.

The joint objective combines text, recognition, and mask losses with an auxiliary spatial-attention loss $\mathcal L_{\mathrm{att}}$, which supervises the segmentation token's attention for non-empty requests. With subset-averaged recognition and mask losses, the joint objective is
\begin{equation}
\label{eq:method-13}
\mathcal L_{\mathrm{joint}}=\mathcal L_{\mathrm{text}}+
\frac{n_r}{B}\mathcal L_{\mathrm{rec}}+
\frac{n_s}{B}\bigl(\mathcal L_{\mathrm{mask}}+
\lambda_a\mathcal L_{\mathrm{att}}\bigr).
\end{equation}
The corresponding term is omitted when its subset is empty. The subset fractions preserve the sample weighting of the mixed batch.

During nine-task training, the Qwen visual encoder and shared Swin-B are frozen. The language model is adapted through LoRA on query and value projections; its output head, temporal fusion, segmentation-token projection, pixel and mask prediction modules, and recognition heads are trainable. A segmentation-enriched warmup stage precedes the main joint-training phase. All nine tasks remain in the sampling pool during these phases, and the optimizer state is carried over between them. Module-specific learning rates and the task sampling configuration are implementation settings for this joint objective.

Both DisasterNexus-4B and DisasterNexus-8B use a single checkpoint, obtained after 18,386 training updates, to evaluate all nine tasks.

\section{Experiments}
\label{sec:experiments}

\subsection{Experimental Setup}

\label{sec:experimental-setup}

We train on DisasterM3 Instruct (87,387 training records after an event-based split of referring segmentation) and evaluate six short-answer tasks, referring segmentation, disaster-report generation, and restoration-advice generation on the released DisasterM3 Bench (30,026 records in total) \cite{wang2025disasterm3}. We report optical--optical and optical--SAR results separately; dashes in Table~\ref{tab:qa-results} indicate that the latter has no DSR records.

Short-answer tasks use exact match and task-level macro-averaging. Segmentation uses cIoU and mIoU, supplemented by non-empty overlap and empty-target accuracy; the main mask threshold is 0.3. Long-form outputs are scored by Qwen3-32B-AWQ using the DisasterM3 rubric, averaging DAP/DDR/FC for reports and RN/APP/SC for advice (excluding IF). Both means are computed from unrounded component scores. The three metric families are reported separately.

The primary model is Qwen3-VL-4B-based DisasterNexus with a frozen Swin-B spatial encoder and rank-8 language LoRA ($\alpha=16$, dropout $0.05$ on $q_{proj}$ and $v_{proj}$). The semantic and spatial paths receive $512\times512$ and $1024\times1024$ images, respectively, with 32 mask-decoder queries. Training runs for 18,386 updates in total with an effective batch size of 16 using AdamW (bfloat16, weight decay 0.01, gradient clipping 1.0). During the 1,000-step warmup prefix, learning rates are $2\times10^{-5}$ for LoRA, $1\times10^{-5}$ for the LM head, and $1\times10^{-4}$ for segmentation and option heads; in the main joint phase, they are $1\times10^{-5}$, $5\times10^{-6}$, and $1\times10^{-4}$, respectively; the final 1,000-step long-form continuation updates only language LoRA ($1\times10^{-5}$) and the LM head ($5\times10^{-6}$) with a cosine schedule. Loss coefficients are $\lambda_b=\lambda_d=5$, $\lambda_q=1$, selector temperature $T=16$, and $\lambda_a=0.01$. Tables~\ref{tab:qa-results} and \ref{tab:refseg-published} mark results from the DisasterM3 paper with $\dagger$; FT denotes fine-tuning by the original authors \cite{wang2025disasterm3}. DisasterNexus results are evaluated locally on Bench, and all models in Table~\ref{tab:longform-comparison} are run and scored locally on the same records.

\subsection{Main Comparative Results}

Tables~\ref{tab:qa-results}--\ref{tab:longform-comparison} report short-answer, referring-segmentation, and long-form generation results, evaluating semantic judgments, spatial localization, and disaster descriptions within one model.

\begin{figure*}[!t]
\centering
\includegraphics[width=\textwidth,height=0.75\textheight,keepaspectratio]{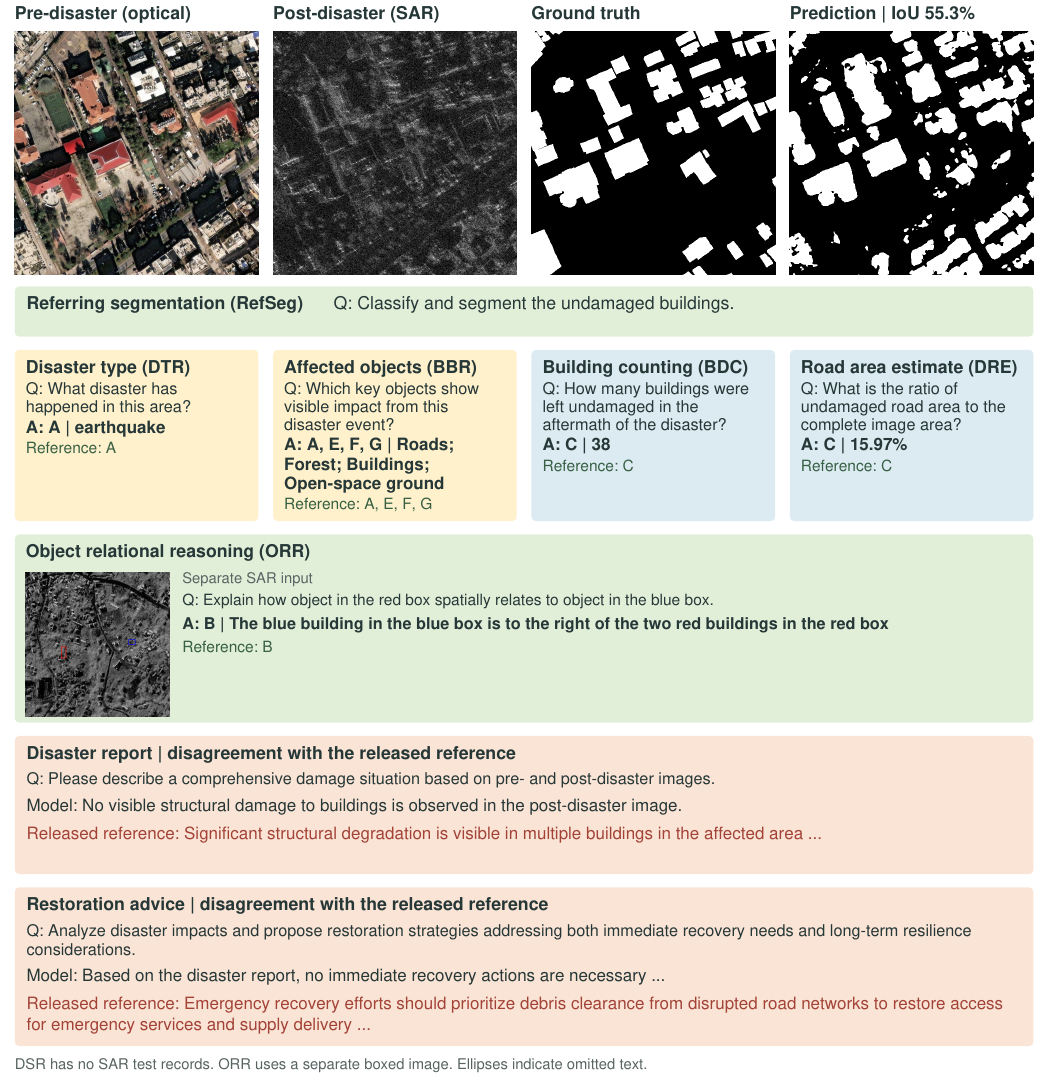}
\caption{DisasterNexus predictions for an optical--SAR earthquake case. ORR uses a single SAR image with its original bounding boxes. Red text quotes the released benchmark reference where it differs from the model output.}
\label{fig:sar-predictions}
\end{figure*}

\begin{figure*}[!t]
\centering
\includegraphics[width=\textwidth,height=0.72\textheight,keepaspectratio]{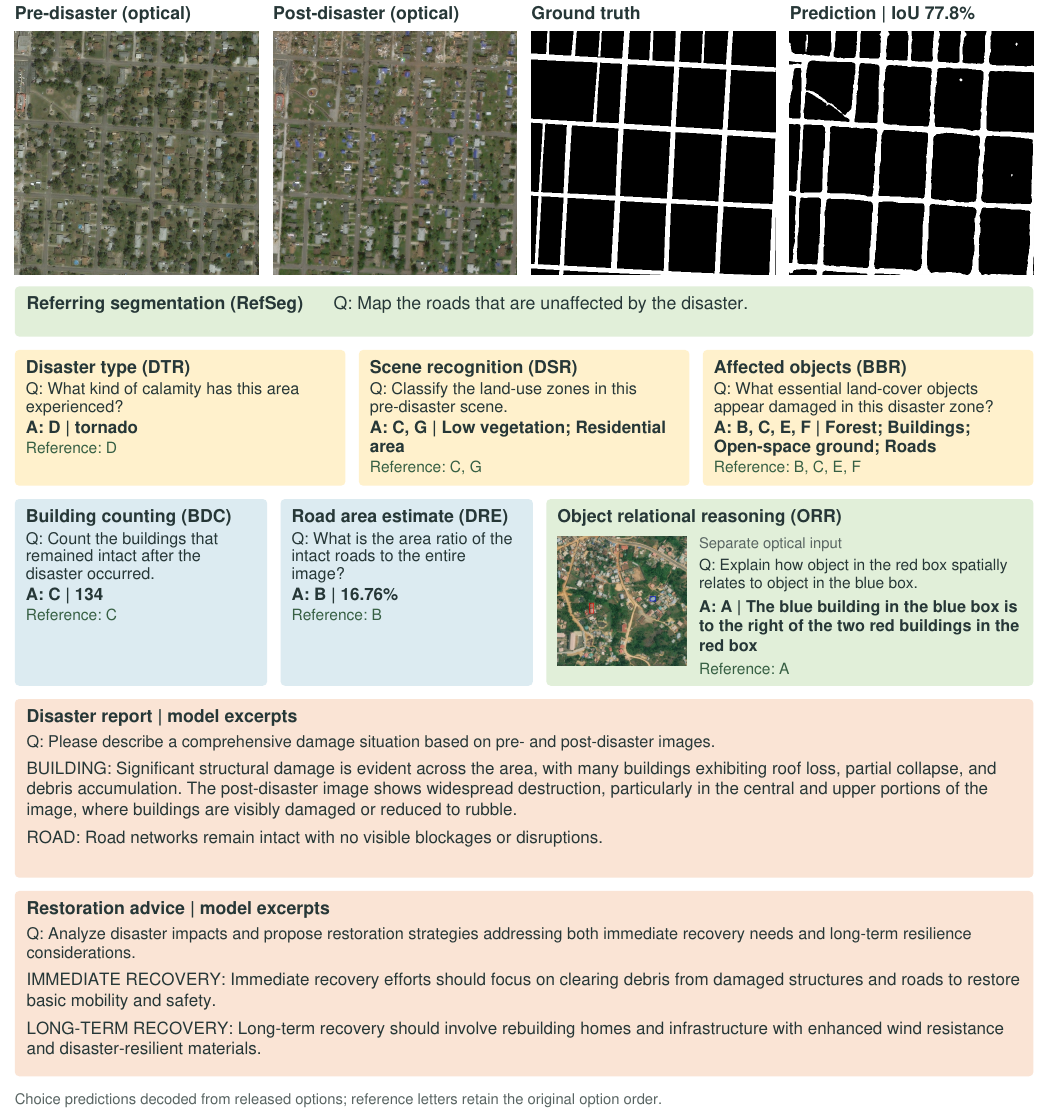}
\caption{Representative nine-task predictions from DisasterNexus on an optical--optical case (Joplin tornado). ORR uses a separate released optical image with the original bounding boxes.}
\label{fig:nine-task-predictions}
\end{figure*}

Complementing the optical--SAR case in Fig.~\ref{fig:sar-predictions}, Fig.~\ref{fig:nine-task-predictions} illustrates unified nine-task predictions on an optical--optical tornado event, showing consistent recognition, damage counting, pixel-level mask localization, and structured report and advice generation from a single checkpoint.

The 4B and 8B models achieve six-task macro-averaged accuracies of 72.58\% and 73.95\%, respectively. The 8B model scores higher on disaster scene and type recognition, optical building counting, and road-damage quantification, whereas the 4B model remains higher on disaster-bearing-body recognition, indicating that each model scale has strengths on different tasks.

\begin{table*}[!t]
\centering
\footnotesize
\setlength{\tabcolsep}{2.5pt}
\renewcommand{\arraystretch}{1.08}
\caption{Short-answer disaster understanding: exact-match accuracy (\%).}
\label{tab:qa-results}
\begin{tabularx}{\textwidth}{@{}>{\raggedright\arraybackslash}Xrrrrrrr@{}}
\toprule
Model & DSR & DTR & BBR & BDC & DRE & ORR & Mean \\
\midrule
\multicolumn{8}{l}{\textit{(a) Optical--optical (six-task mean)}} \\
Qwen2.5-VL-7B$^{\dagger}$ & 28.30 & 66.60 & 4.70 & 34.20 & 29.30 & 23.90 & 31.17 \\
InternVL3-8B$^{\dagger}$ & 39.60 & 53.50 & 4.00 & 30.30 & 24.10 & 36.20 & 31.28 \\
TeoChat-7B$^{\dagger}$ & 6.90 & 64.90 & 2.00 & 22.50 & 23.30 & 18.20 & 22.97 \\
EarthDial-4B$^{\dagger}$ & 10.60 & 58.10 & 3.20 & 30.20 & 20.80 & 14.50 & 22.90 \\
Qwen2.5-VL-7B-FT$^{\dagger}$ & 37.70 & 83.60 & 21.50 & 34.30 & 29.40 & 36.20 & 40.45 \\
InternVL3-8B-FT$^{\dagger}$ & 42.60 & 79.30 & 23.90 & 29.10 & 24.90 & 50.60 & 41.73 \\
\textbf{DisasterNexus-4B} & 55.33 & 84.42 & \textbf{59.09} & 73.92 & 62.02 & 97.90 & 72.11 \\
\textbf{DisasterNexus-8B} & \textbf{58.37} & \textbf{87.54} & 55.84 & \textbf{77.77} & \textbf{64.60} & \textbf{98.37} & \textbf{73.75} \\
\midrule
\multicolumn{8}{l}{\textit{(b) Optical--SAR (five-task mean)}} \\
Qwen2.5-VL-7B$^{\dagger}$ & -- & 62.70 & 8.40 & 16.90 & 11.90 & 13.10 & 22.60 \\
InternVL3-8B$^{\dagger}$ & -- & 32.80 & 7.30 & 20.70 & 18.40 & 28.10 & 21.46 \\
TeoChat-7B$^{\dagger}$ & -- & 29.90 & 4.50 & 18.40 & 9.40 & 13.10 & 15.06 \\
EarthDial-4B$^{\dagger}$ & -- & 30.70 & 6.80 & 19.50 & 10.20 & 14.30 & 16.30 \\
Qwen2.5-VL-7B-FT$^{\dagger}$ & -- & 64.20 & 21.00 & 29.40 & 13.90 & 21.20 & 29.94 \\
InternVL3-8B-FT$^{\dagger}$ & -- & 73.10 & 18.80 & 23.60 & 18.70 & 36.20 & 34.08 \\
\textbf{DisasterNexus-4B} & -- & 89.55 & \textbf{55.34} & \textbf{75.70} & 78.71 & \textbf{95.62} & 78.98 \\
\textbf{DisasterNexus-8B} & -- & \textbf{92.54} & 50.56 & 75.41 & \textbf{81.61} & 95.00 & \textbf{79.02} \\
\bottomrule
\end{tabularx}
\end{table*}

The 4B and 8B models show comparable pixel-localization performance, with pooled cIoU of 49.06\% and 49.19\%, respectively (Table~\ref{tab:refseg-published}). Section~\ref{sec:grouped-errors} compares DisasterNexus with LISA and PSALM locally on the same set of non-empty requests.

\begin{table*}[!t]
\centering
\footnotesize
\setlength{\tabcolsep}{2.5pt}
\renewcommand{\arraystretch}{1.08}
\caption{Disaster referring segmentation: mIoU and cIoU (\%).}
\label{tab:refseg-published}
\begin{tabularx}{\textwidth}{@{}>{\raggedright\arraybackslash}Xrrrr@{}}
\toprule
Model & \shortstack{Opt.--Opt.\\mIoU} & \shortstack{Opt.--Opt.\\cIoU} & \shortstack{Opt.--SAR\\mIoU} & \shortstack{Opt.--SAR\\cIoU} \\
\midrule
\multicolumn{5}{l}{\textit{Reported references}} \\
LISA-7B$^{\dagger}$ & 27.50 & 22.10 & 10.90 & 12.70 \\
PSALM-1.3B$^{\dagger}$ & 9.70 & 6.30 & 8.10 & 8.80 \\
HyperSeg-3B$^{\dagger}$ & 16.60 & 14.50 & 8.80 & 10.30 \\
GeoPixel-7B$^{\dagger}$ & 14.30 & 14.20 & 4.00 & 5.10 \\
LISA-7B-FT$^{\dagger}$ & 44.80 & 43.70 & 28.20 & 29.60 \\
PSALM-1.3B-FT$^{\dagger}$ & 50.50 & 44.50 & 31.80 & 35.20 \\
\midrule
\multicolumn{5}{l}{\textit{Local comparisons}} \\
\textbf{DisasterNexus-4B} & 61.47 & \textbf{50.81} & \textbf{49.72} & 44.85 \\
\textbf{DisasterNexus-8B} & \textbf{61.52} & 50.68 & 49.34 & \textbf{45.57} \\
\bottomrule
\end{tabularx}
\end{table*}

\begin{figure*}[!t]
\centering
\includegraphics[width=\textwidth]{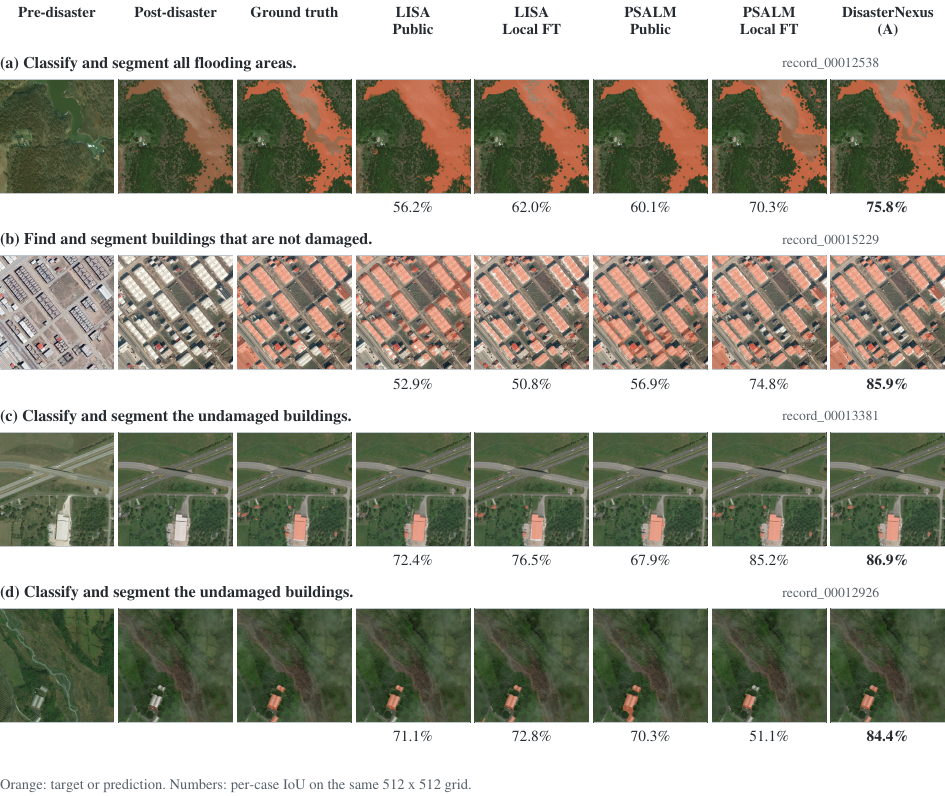}
\caption{Selected optical segmentation comparisons on identical source images and instructions. Columns show pre-event and post-event images, ground truth, LISA public/local FT, PSALM public/local FT, and DisasterNexus. Local FT denotes our own adaptation, distinct from the published FT rows in Table~\ref{tab:refseg-published}. LISA/PSALM use a left--right paired-image composite of pre- and post-event images. Native binary predictions are retained; nearest-neighbor resizing maps masks to a common 512$\times$512 grid for the displayed IoU. Orange marks target/predicted regions.}
\label{fig:segmentation-method-comparison}
\end{figure*}

DisasterNexus-4B achieves overall scores of 3.450 for disaster reports and 2.734 for recovery advice, while DisasterNexus-8B reaches 3.546 and 2.844, both exceeding the highest scores among the evaluated public checkpoints for each task (2.320 and 2.408, respectively). On the factual correctness criterion for reports on optical imagery, TeoChat scores higher than both DisasterNexus variants. The independent two-task model scores 3.493 for reports and 2.940 for advice; DisasterNexus-8B surpasses it on disaster reports while trailing on recovery advice (Table~\ref{tab:longform-comparison}).

\begin{table*}[!t]
\centering
\footnotesize
\setlength{\tabcolsep}{2.5pt}
\renewcommand{\arraystretch}{1.08}
\caption{Local evaluation of public checkpoints and trained models for disaster report and restoration-advice generation using Qwen3-32B-AWQ (0--5).}
\label{tab:longform-comparison}
\begin{tabularx}{\textwidth}{@{}>{\raggedright\arraybackslash}Xrrrrrrrr@{}}
\toprule
Model & DAP & DDR & FC & \shortstack{Report\\mean} & RN & APP & SC & \shortstack{Advice\\mean} \\
\midrule
\multicolumn{9}{l}{\textit{(a) Optical--optical}} \\
Qwen2.5-VL-7B & 1.88 & 1.26 & 2.23 & 1.79 & 3.26 & 2.19 & 1.95 & 2.47 \\
InternVL3-8B & 1.55 & 1.21 & 1.77 & 1.51 & 3.11 & 2.11 & 1.75 & 2.32 \\
TeoChat-7B & 2.07 & 1.54 & \textbf{3.96} & 2.52 & 3.51 & 1.94 & 1.75 & 2.40 \\
EarthDial-4B & 1.51 & 0.64 & 2.24 & 1.47 & 2.69 & 1.56 & 1.16 & 1.80 \\
\addlinespace[2pt]
\textbf{DisasterNexus-4B} & 3.48 & 3.24 & 3.72 & 3.48 & 3.44 & 2.38 & 2.34 & 2.72 \\
\textbf{DisasterNexus-8B} & \textbf{3.57} & \textbf{3.34} & 3.86 & \textbf{3.59} & 3.58 & 2.50 & 2.45 & 2.84 \\
Qwen3-VL-4B (two-task SFT) & 3.53 & 3.28 & 3.82 & 3.54 & \textbf{3.73} & \textbf{2.60} & \textbf{2.51} & \textbf{2.95} \\
\midrule
\multicolumn{9}{l}{\textit{(b) Optical--SAR}} \\
Qwen2.5-VL-7B & 1.41 & 0.87 & 1.84 & 1.37 & 2.83 & 1.76 & 1.62 & 2.07 \\
InternVL3-8B & 1.24 & 0.84 & 1.59 & 1.22 & 2.70 & 1.86 & 1.54 & 2.03 \\
TeoChat-7B & 0.67 & 0.33 & 2.50 & 1.17 & 2.96 & 1.66 & 1.51 & 2.04 \\
EarthDial-4B & 1.18 & 0.59 & 1.72 & 1.16 & 2.00 & 1.08 & 0.73 & 1.27 \\
\addlinespace[2pt]
\textbf{DisasterNexus-4B} & 3.22 & 3.01 & 3.59 & 3.27 & 3.79 & 2.29 & 2.31 & 2.80 \\
\textbf{DisasterNexus-8B} & \textbf{3.24} & \textbf{3.02} & \textbf{3.64} & \textbf{3.30} & 3.59 & \textbf{2.48} & \textbf{2.51} & 2.86 \\
Qwen3-VL-4B (two-task SFT) & 3.17 & 2.97 & 3.51 & 3.22 & \textbf{3.96} & 2.37 & 2.36 & \textbf{2.90} \\
\bottomrule
\end{tabularx}
\end{table*}

\begin{figure*}[!t]
\centering
\includegraphics[width=\textwidth,height=0.75\textheight,keepaspectratio]{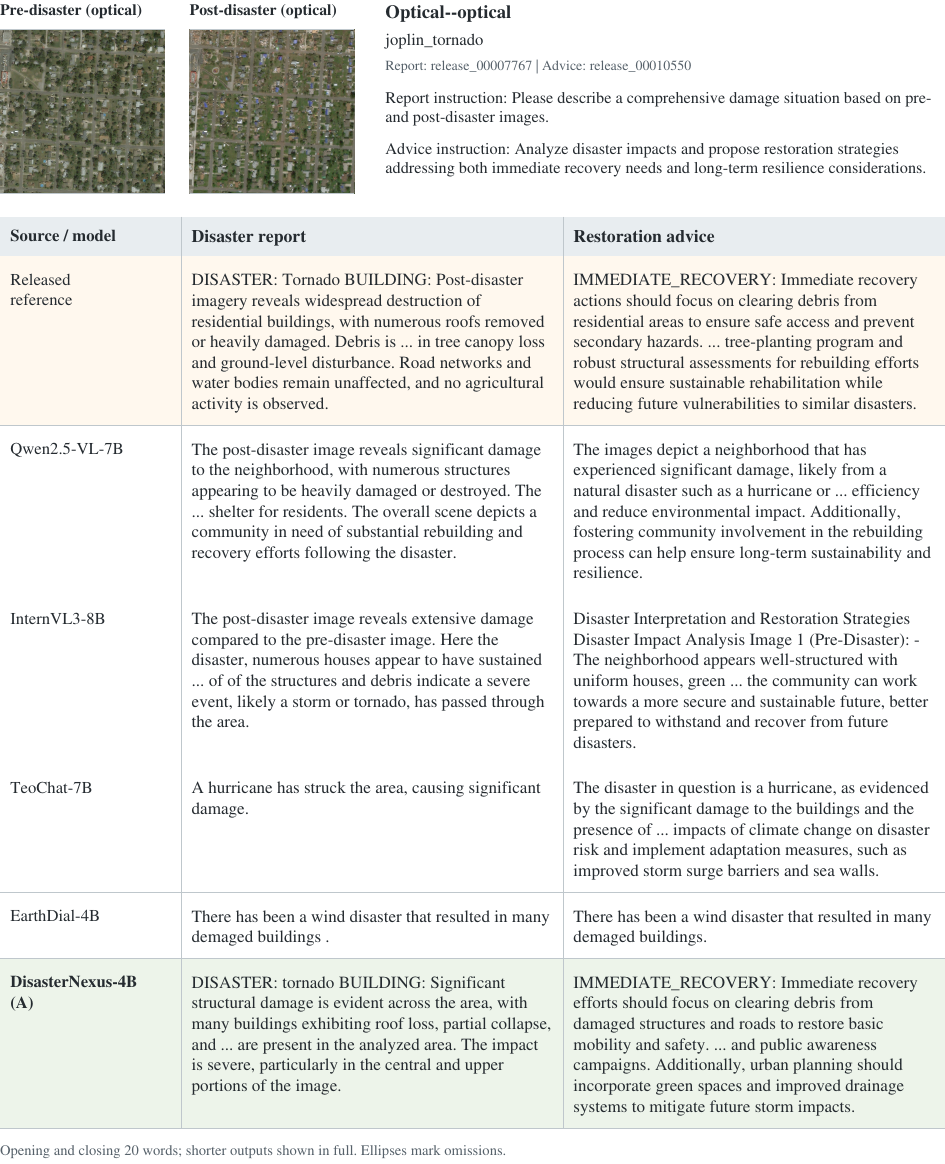}
\caption{Comparison of disaster reports and recovery advice for the same optical image pair. The figure shows the benchmark reference and excerpts from the original model outputs. Scores in Table~\ref{tab:longform-comparison} use the full outputs.}
\label{fig:longform-method-comparison}
\end{figure*}

Figure~\ref{fig:longform-method-comparison} compares disaster reports and recovery advice generated by the models for the same optical case. DisasterNexus identifies the event as a tornado and reports roof loss and partial building collapse. Its recovery advice proposes immediate measures such as clearing debris and restoring access. TeoChat describes the event as a hurricane, while EarthDial mainly describes wind damage in both responses.

\begin{figure*}[!t]
\centering
\includegraphics[width=\textwidth,height=0.72\textheight,keepaspectratio]{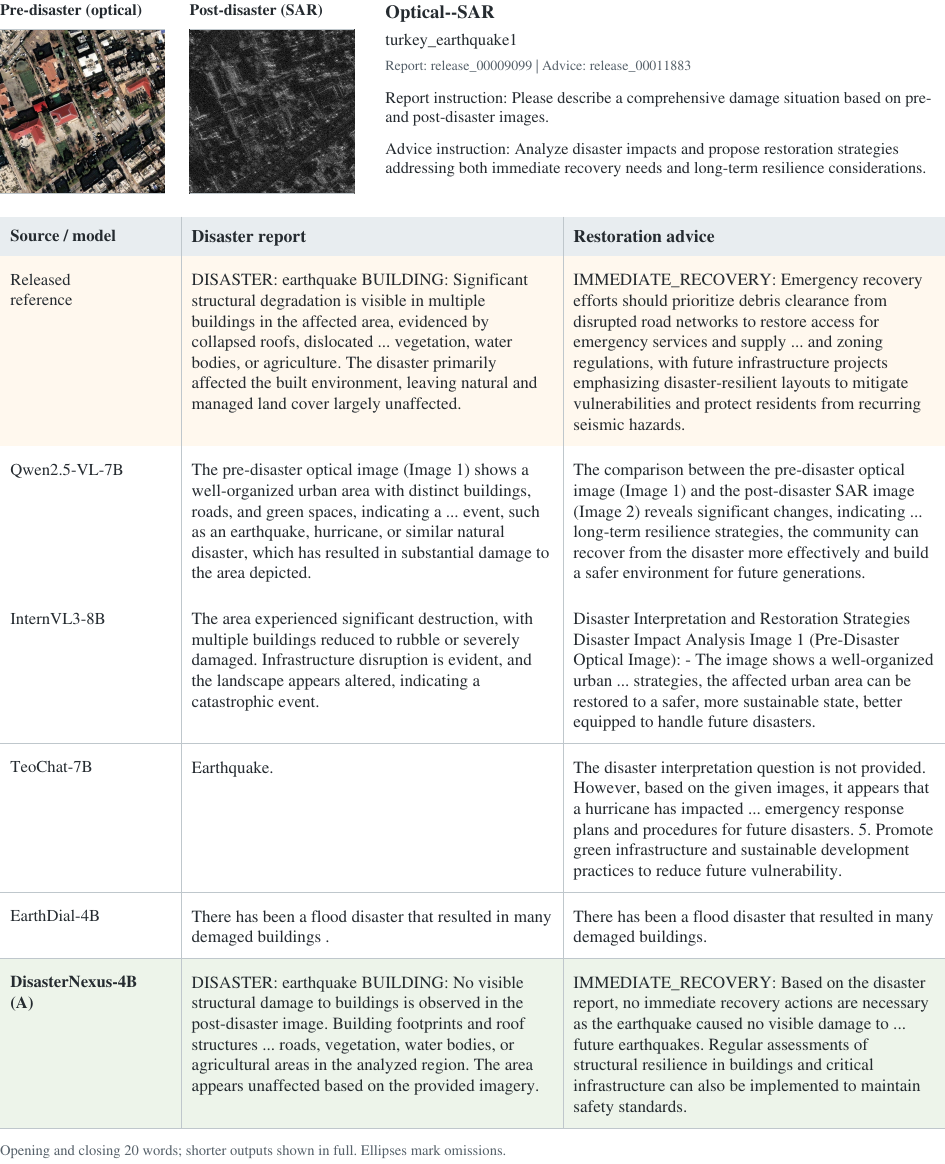}
\caption{Comparison of disaster reports and recovery advice on an optical--SAR image pair. Each cell shows the opening and closing excerpts; scores in Table~\ref{tab:longform-comparison} evaluate the full outputs.}
\label{fig:sar-longform-method-comparison}
\end{figure*}

To evaluate qualitative generation under cross-sensor observations, Fig.~\ref{fig:sar-longform-method-comparison} compares outputs on an optical--SAR pair. While public checkpoints frequently misidentify the disaster context or omit critical infrastructure damage under SAR appearance shifts, DisasterNexus produces accurate damage summaries and actionable restoration recommendations aligned with the benchmark reference.

\subsection{Ablation Studies}
\label{sec:representation-evidence}

\begin{figure*}[!t]
\centering
\includegraphics[width=\textwidth,height=0.74\textheight,keepaspectratio]{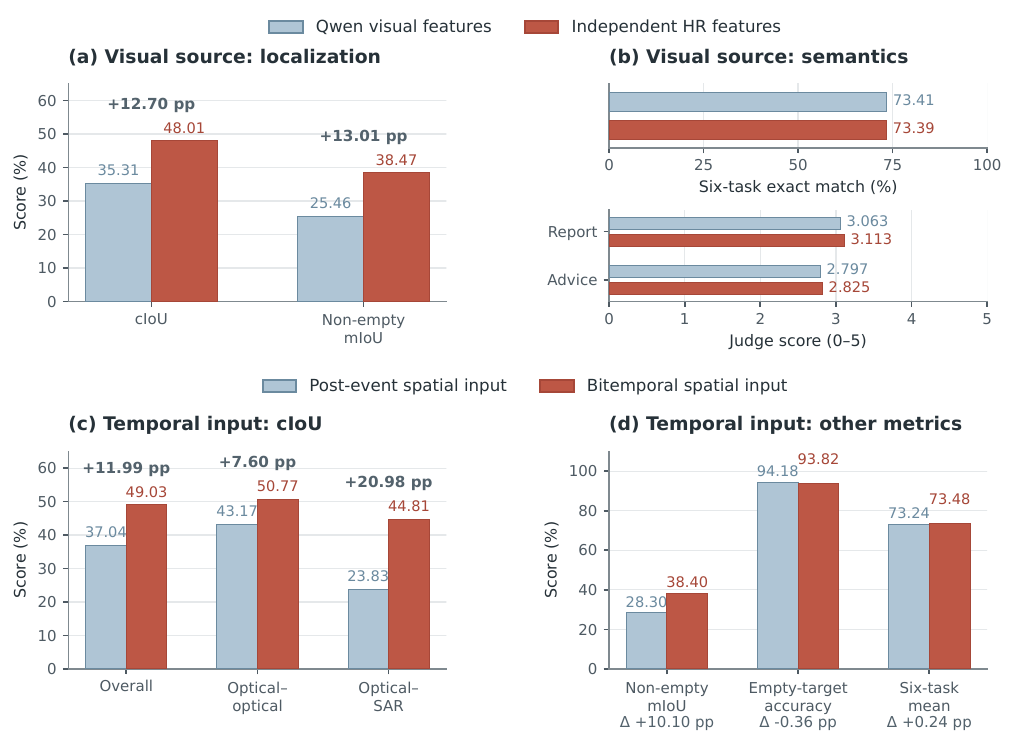}
\caption{Spatial representation and temporal input comparisons after 17,386 joint-training updates.}
\label{fig:ablation-comparison}
\end{figure*}

Figure~\ref{fig:ablation-comparison} presents two independent comparisons. Compared with the Qwen-feature configuration, the independent high-resolution spatial configuration achieves cIoU and non-empty mIoU scores that are 12.70 and 13.01 percentage points higher, respectively, with comparable short-answer and long-form performance (Figure~\ref{fig:ablation-comparison}a,b).

Using both pre- and post-event images in the spatial path yields a larger localization gain for optical--SAR observations (Figure~\ref{fig:ablation-comparison}c). Compared with the post-event-only spatial configuration, the bitemporal spatial configuration raises cIoU from 23.83\% to 44.81\% for optical--SAR pairs, a gain of 20.98 percentage points, and improves cIoU by 7.60 percentage points for optical--optical pairs. The semantic path receives both images in both configurations, and short-answer performance remains comparable. Empty-target accuracy is slightly lower in the bitemporal spatial configuration (Figure~\ref{fig:ablation-comparison}d).

\subsection{Grouped Segmentation Error Comparison}
\label{sec:grouped-errors}

The local comparison covers all 7,684 non-empty DisasterM3 Bench requests, including 5,933 optical--optical and 1,751 optical--SAR requests. We group requests by the fraction of the image covered by the union of ground-truth regions for each instruction: small ($\leq1\%$), medium ($(1\%,10\%]$), and large ($>10\%$), with 2,435, 3,063, and 2,186 requests, respectively.

\begin{figure*}[!t]
\centering
\includegraphics[width=\textwidth]{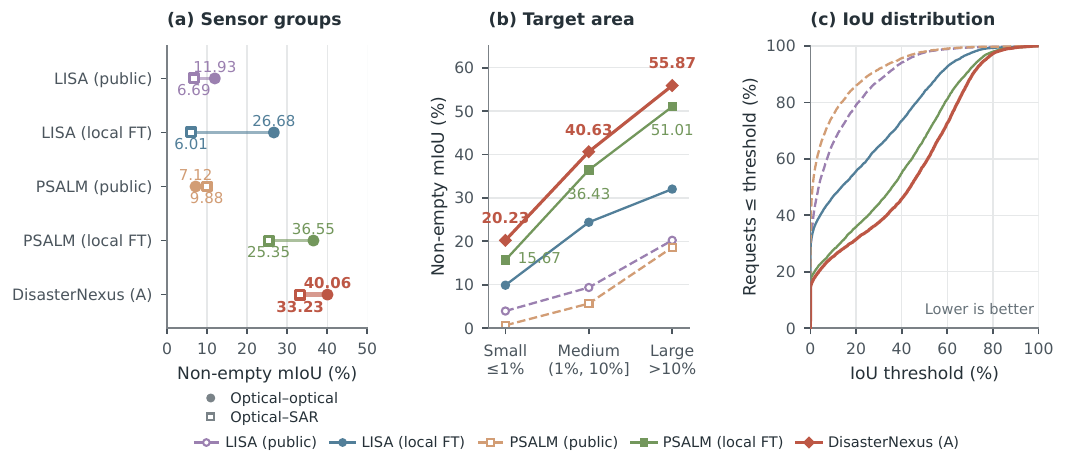}
\caption{Grouped segmentation performance and IoU distribution.}
\label{fig:grouped-segmentation-analysis}
\end{figure*}

DisasterNexus-4B achieves an overall non-empty mIoU of 38.50\%, exceeding locally fine-tuned PSALM by 4.50 percentage points and by 7.89 points on optical--SAR inputs. Adaptation gains differ by sensor: LISA improves on optical--optical inputs but not on optical--SAR (Figure~\ref{fig:grouped-segmentation-analysis}a). All methods perform worst on small-area requests, indicating that fine-scale region localization remains a shared bottleneck (Figure~\ref{fig:grouped-segmentation-analysis}b).

\begin{table*}[!t]
\centering
\footnotesize
\setlength{\tabcolsep}{3pt}
\caption{Grouped segmentation comparison on DisasterM3 Bench (non-empty mIoU, \%).}
\label{tab:grouped-segmentation}
\begin{tabularx}{\textwidth}{@{}>{\raggedright\arraybackslash}Xrrrrrr@{}}
\toprule
Method & Overall & \shortstack{Optical--\\optical} & \shortstack{Optical--\\SAR} & Small & Medium & Large \\
\midrule
LISA (public) & 10.73 & 11.93 & 6.69 & 3.94 & 9.36 & 20.23 \\
LISA (local FT) & 21.97 & 26.68 & 6.01 & 9.89 & 24.39 & 32.03 \\
PSALM (public) & 7.75 & 7.12 & 9.88 & 0.64 & 5.66 & 18.60 \\
PSALM (local FT) & 34.00 & 36.55 & 25.35 & 15.67 & 36.43 & 51.01 \\
\textbf{DisasterNexus-4B} & \textbf{38.50} & \textbf{40.06} & \textbf{33.23} & \textbf{20.23} & \textbf{40.63} & \textbf{55.87} \\
\bottomrule
\end{tabularx}
\end{table*}

Figure~\ref{fig:grouped-segmentation-analysis}(c) shows the empirical cumulative distribution of IoU across all non-empty requests, including predictions with zero IoU. The fraction of requests with IoU above 0.5 reaches 41.91\% for DisasterNexus-4B, compared with 32.65\% for locally fine-tuned PSALM.

\subsection{Multi-Task Synergy and Target-Presence Analysis}
\label{sec:synergy-presence}

To verify whether unifying nine tasks compromises dense pixel localization, Table~\ref{tab:synergy-presence} compares DisasterNexus against a single-encoder nine-task baseline (Basic-HR) and a dedicated single-task referring-segmentation model trained with comparable segmentation supervision exposures (63,246 vs.~63,508). While Basic-HR drops to 25.35\% cIoU under multi-task interference, DisasterNexus-4B and DisasterNexus-8B achieve 49.06\% and 49.19\% cIoU, outperforming Basic-HR by +23.71 and +23.84 percentage points, respectively, and approaching the dedicated single-task ceiling (50.02\% cIoU, within 0.83--0.96 points) while simultaneously supporting eight additional semantic and generative tasks.

\begin{table}[!t]
\centering
\footnotesize
\setlength{\tabcolsep}{3pt}
\renewcommand{\arraystretch}{1.08}
\caption{Multi-task synergy and target-presence metrics on DisasterM3 Bench (\%, mask threshold 0.3).}
\label{tab:synergy-presence}
\begin{tabularx}{\columnwidth}{@{}>{\raggedright\arraybackslash}Xrrrr@{}}
\toprule
Model / split & mIoU & cIoU & \shortstack{Non-empty\\mIoU} & \shortstack{Empty\\acc.} \\
\midrule
\multicolumn{5}{l}{\textit{(a) Nine-task vs.~single-task (all requests)}} \\
Basic-HR (nine tasks) & 48.91 & 25.35 & 21.85 & 93.54 \\
RefSeg single-task & \textbf{60.16} & \textbf{50.02} & \textbf{39.58} & \textbf{94.10} \\
\textbf{DisasterNexus-4B} & 59.25 & 49.06 & 38.50 & 93.47 \\
\textbf{DisasterNexus-8B} & 59.22 & 49.19 & 38.13 & 94.01 \\
\midrule
\multicolumn{5}{l}{\textit{(b) Modality breakdown for DisasterNexus}} \\
4B (Opt.--Opt.) & 61.47 & \textbf{50.81} & \textbf{40.06} & 92.60 \\
8B (Opt.--Opt.) & \textbf{61.52} & 50.68 & 39.72 & 93.21 \\
4B (Opt.--SAR) & 49.72 & 44.85 & 33.23 & \textbf{99.65} \\
8B (Opt.--SAR) & 49.34 & 45.57 & 32.73 & \textbf{99.65} \\
\bottomrule
\end{tabularx}
\end{table}

Table~\ref{tab:synergy-presence}(b) further highlights strong target-presence discrimination across sensors. Overall empty-target accuracy reaches 93.47\% for DisasterNexus-4B and 94.01\% for DisasterNexus-8B, rising to 99.65\% on optical--SAR pairs, confirming that the model reliably rejects unaffected scenes without producing false-positive masks. Moreover, candidate-pool diagnostics on the 32 mask queries show an average of 27.25 distinct candidates (under an IoU-0.9 duplicate criterion), verifying that joint multi-task training preserves diverse spatial proposals without query collapse.

\section{Conclusion}
\label{sec:conclusion}

We have presented DisasterNexus to address a central challenge in remote-sensing disaster analysis: enabling a single model to perform scene-level semantic interpretation alongside pixel-level localization of instruction-specified regions. By pairing a contextual vision--language branch with an independent high-resolution bitemporal spatial branch, a single 4B checkpoint handles all nine DisasterM3 tasks within one unified framework. Under matched training data and budgets, our independent spatial branch yields a 12.70-percentage-point gain in referring segmentation cIoU over the Qwen-feature baseline without degrading short-answer accuracy.

These findings confirm that dedicating a high-resolution spatial pathway to mask generation---while leaving candidate selection to the language model's semantic representation---offers an effective division of labor for unified disaster understanding. Nevertheless, small-area targets and cross-sensor alignment remain key bottlenecks. Future work will focus on scale-aware candidate generation for small objects and stronger optical--SAR feature alignment to further improve localization accuracy in complex disaster scenes.


\end{document}